\documentclass{article}

\usepackage{arxiv}

\usepackage[utf8]{inputenc} % allow utf-8 input
\usepackage[T1]{fontenc}    % use 8-bit T1 fonts
\usepackage{hyperref}       % hyperlinks
\usepackage{url}            % simple URL typesetting
\usepackage{booktabs}       % professional-quality tables
\usepackage{amsfonts}       % blackboard math symbols
\usepackage{nicefrac}       % compact symbols for 1/2, etc.
\usepackage{microtype}      % microtypography
\usepackage{lipsum}
\usepackage{graphicx}
\graphicspath{ {./images/} }
\usepackage{algorithm}
\usepackage{algpseudocode}  % 替代 algorithmic
\usepackage{booktabs}
\usepackage{multirow}
\usepackage{amsmath}        % 核心数学包
\usepackage{amssymb}        % 更多数学符号
\usepackage{mathtools}      % amsmath 的增强版
\usepackage{bm}             % 加粗数学符号
\usepackage{physics}        % 物理相关符号（可选）

\title{Agent Guide: A Simple Agent Behavioral Watermarking Framework}

\author{
 Kaibo Huang \\
  Beijing University of Posts and Telecommunications \\
  Beijing, China \\
  \texttt{Huangkaibo@bupt.edu.cn} \\
  \And
  Zipei Zhang \\
  Beijing University of Posts and Telecommunications \\
  Beijing, China \\
  \texttt{nebulazhang@bupt.edu.cn} \\
  \And
 Zhongliang Yang\thanks{Corresponding author.} \\
  Beijing University of Posts and Telecommunications \\
  Beijing, China \\
  \texttt{yangzl@bupt.edu.cn} \\
  \And
 Linna Zhou \\
  Beijing University of Posts and Telecommunications \\
  Beijing, China \\
  \texttt{zhoulinna@bupt.edu.cn} \\
}

\begin{document}
\maketitle

\begin{center}
\textbf{Code:} \url{https://github.com/Tooooa/AgentMark}\footnote{The implementation of Agent Guide is included as a baseline in the AgentMark repository.}
\end{center}

\begin{abstract}
The increasing deployment of intelligent agents in digital ecosystems, such as social media platforms, has raised significant concerns about traceability and accountability, particularly in cybersecurity and digital content protection. Traditional large language model (LLM) watermarking techniques, which rely on token-level manipulations, are ill-suited for agents due to the challenges of behavior tokenization and information loss during behavior-to-action translation. To address these issues, we propose \textit{Agent Guide}, a novel behavioral watermarking framework that embeds watermarks by guiding the agent's high-level decisions (\textbf{behavior}) through probability biases, while preserving the naturalness of specific executions (\textbf{action}). Our approach decouples agent behavior into two levels—\textbf{behavior} (e.g., choosing to bookmark) and \textbf{action} (e.g., bookmarking with specific tags)—and applies watermark-guided biases to the behavior probability distribution. We employ a z-statistic-based statistical analysis to detect the watermark, ensuring reliable extraction over multiple rounds. Experiments in a social media scenario with diverse agent profiles demonstrate that \textit{Agent Guide} achieves effective watermark detection with a low false positive rate. Our framework provides a practical and robust solution for agent watermarking, with applications in identifying malicious agents and protecting proprietary agent systems.
\end{abstract}

% keywords can be removed

\section{Introduction}
\label{sec:introduction}
In recent years, large language models (LLMs) have emerged as a transformative technology, leveraging vast knowledge encoded from diverse web data to understand and generate human-like content~\cite{brown2020language, mei2024turing}. This remarkable capability has positioned LLMs as powerful tools for simulating human interactions, particularly in scenarios requiring subjective decision-making, dynamic interaction patterns, and personalized preferences~\cite{hadi2023survey}. As a result, there has been a growing interest in deploying LLM-powered agents to perform complex tasks across various domains, ranging from virtual assistants~\cite{hardy2023large} to automated content generation systems~\cite{hong2023metagpt}. These agents are increasingly capable of exhibiting human-like behaviors with high fidelity, making their actions nearly indistinguishable from real users in certain contexts~\cite{zhang2023human}.

The proliferation of intelligent agents has opened up significant opportunities while simultaneously introducing critical challenges in the domains of cybersecurity and copyright protection~\cite{deng2025ai}. In recent years, agents have seen rapid advancements in simulating user behavior on social platforms, achieving unprecedented levels of realism in mimicking human interactions~\cite{gao2023s3, wang2025user, park2023generative, piao2025agentsociety}. This capability, however, can be exploited by malicious actors. For instance, adversaries may deploy agents on social platforms to impersonate real users, posting misleading or fraudulent information—such as impersonating a celebrity to promote fake charity campaigns—to manipulate public opinion, promote scams, or spread disinformation~\cite{mou2024unveiling}. The sophisticated conversational capabilities of these agents, combined with the vast reach of social platforms, make such malicious activities difficult to detect and regulate. Beyond cybersecurity, the rise of agents also poses risks to copyright protection, as enterprises invest significant resources in training specialized agents for applications like customer service, content moderation, or creative content generation, which embody proprietary knowledge and behavioral patterns that are valuable intellectual property~\cite{barfield2022considering}. Once deployed, these agents are vulnerable to unauthorized replication or misuse by competitors, leading to potential intellectual property theft. These growing threats underscore the urgent need for effective identification and traceability of agent behaviors in digital ecosystems to ensure accountability, mitigate risks, and safeguard ownership of proprietary agent systems.

To address the challenge of identifying and tracing agent behaviors in digital ecosystems, watermarking has emerged as a promising mechanism for embedding traceable markers into digital entities. In the context of intelligent agents, watermarking can enable the identification of malicious agents in cybersecurity scenarios, such as detecting agents spreading disinformation, or verify ownership in copyright protection scenarios, such as protecting proprietary customer service agents. Existing watermarking techniques have primarily focused on content watermarking, which aims to distinguish whether online content is generated by AI or human users~\cite{kirchenbauer2023watermark, pan2024markllm}. While content watermarking is effective for tracing the origin of LLM-generated outputs, it does not address the broader spectrum of agent activities, particularly their behavioral patterns. For instance, an agent's generated content, such as a misleading post, may cause harm, but its behavior—such as repeatedly sharing such posts to amplify disinformation—can be equally or more damaging due to its intent and scale of impact. 

\textbf{Both content and behavior require traceability and accountability, yet behavior often plays a more critical role in understanding an agent's decision-making and long-term effects, making behavioral watermarking an essential complement to content watermarking.}

However, applying content watermarking techniques to agent behaviors faces significant challenges. First, some methods embed watermarks during the training phase of large language models (LLMs) by modifying model weights~\cite{lau2024waterfall, patil2023can, peng2023you}. This approach is impractical for agent systems, as most agents directly invoke pre-trained LLM APIs (e.g., GPT-4~\cite{achiam2023gpt} or Deepseek~\cite{guo2025deepseek}), making it impossible to intervene in the training process. Moreover, these pre-trained models are general-purpose and not optimized for specific agent scenarios, and retraining them incurs high computational costs while failing to dynamically adjust watermark strategies to meet diverse agent requirements. Second, other methods embed watermarks by manipulating token-level probabilities during logits generation or token sampling~\cite{kirchenbauer2023watermark, dathathri2024scalable, guan2024codeip, hudecek-dusek-2023-large}. These token-centric approaches cannot effectively capture an agent's high-level \textbf{behavior}, such as the decision to bookmark a post, because such behaviors are misaligned with token structures. For instance, a single behavior like ``bookmarking'' might be tokenized into multiple tokens (e.g., ``book'' and ``marking''), making it difficult to align token-level watermark manipulations with the agent's intended behavior. Additionally, when an agent's \textbf{behavior} is translated into discrete \textbf{action}, significant information is lost. For example, an LLM output like ``Alice bookmarked a post with the tag `\texttt{\#TravelInspiration}''' is simplified into actions like ``bookmarking'' and ``tagging,'' stripping away contextual details and intent, which makes embedded watermarks difficult to extract and verify. Unlike traditional content generation, agents operate in a data-and-behavior-integrated paradigm, where high-level decisions (\textbf{behavior}) are translated into specific executions (\textbf{action}). This behavioral layer, which encapsulates the agent's intent and decision-making, offers a robust and semantically rich target for watermarking, often providing deeper insights into an agent's purpose and impact compared to content alone. To the best of our knowledge, there is currently no watermarking technique specifically designed for the network behaviors of AI agents, highlighting the urgent need for behavioral watermarking to ensure comprehensive traceability and accountability in digital ecosystems.

In this paper, we propose \textit{Agent Guide}, a novel behavioral watermarking framework for agents that introduces the first method specifically targeting the network behaviors of AI agents. Our approach embeds watermarks by applying biases to the agent's high-level decisions (\textbf{behavior})—such as choosing to bookmark or deciding to like—while preserving the naturalness of specific executions (\textbf{action}), such as the specific tags used in bookmarking. By decoupling an agent's behavior into these two levels and introducing watermark biases at the behavior level, \textit{Agent Guide} embeds detectable markers into the agent's decision probability distribution, ensuring that the action outputs remain natural and unaffected. Our key contributions are as follows:

\begin{itemize}
    \item We propose \textit{Agent Guide}, the first behavioral watermarking framework specifically designed for the network behaviors of AI agents, addressing the limitations of content watermarking by focusing on high-level behavioral patterns.
    \item Our method enables effective agent identification and copyright protection, providing a practical and scalable solution for applications such as detecting malicious agents in cybersecurity scenarios, safeguarding proprietary agent systems, and ensuring compliance in regulated industries like finance and healthcare.
    \item We validate the effectiveness and robustness of \textit{Agent Guide} through extensive experiments across diverse agent profiles, demonstrating its ability to achieve reliable watermark detection with low false positive rates in real-world scenarios.
\end{itemize}

\section{Related Work}

\subsection{Advances in Intelligent Agents}
The rapid development of large language models (LLMs) has significantly advanced the capabilities of intelligent agents, enabling them to perform complex tasks across various domains~\cite{brown2020language, mei2024turing}. A key area of progress is agent-based platform simulation, where agents are deployed to mimic human interactions on social platforms with high fidelity~\cite{gao2023s3, wang2025user, park2023generative, piao2025agentsociety}. For instance, Gao \emph{et al.}~\cite{gao2023s3} proposed a framework for simulating social media interactions, allowing agents to generate realistic user behaviors such as posting, commenting, and sharing. Similarly, Piao \emph{et al.}~\cite{piao2025agentsociety} developed an agent society model that replicates dynamic group interactions, achieving unprecedented levels of realism. Another important direction is the use of agents to assist humans in decision-making and task automation~\cite{hong2023metagpt, yu2024finmem}. Hong \emph{et al.}~\cite{hong2023metagpt} explored agents for automated content generation, such as coding and design, by introducing human workflows and standardized operating procedures (SOPs) into multi-agent systems to address logical inconsistencies and hallucination issues in complex tasks. In the financial domain, Yu \emph{et al.}~\cite{yu2024finmem} designed FINMEM, an intelligent agent that assists users in making optimal investment decisions in volatile financial markets by leveraging self-learning and real-time adaptation to enhance trading performance. These advances highlight the growing role of agents in digital ecosystems, particularly in the domains of cybersecurity and copyright protection. In cybersecurity, agents can be exploited to spread disinformation or manipulate public opinion, posing significant risks to digital safety~\cite{mou2024unveiling}. In copyright protection, agents developed for specialized applications, such as customer service or content generation, embody valuable intellectual property that is vulnerable to unauthorized replication or misuse~\cite{barfield2022considering}. These challenges underscore the importance of developing effective mechanisms to ensure the traceability and accountability of agent activities in digital ecosystems.

\subsection{Watermarking Techniques for Agent}
Watermarking has been widely studied as a mechanism to embed traceable markers into LLM outputs, primarily to ensure authenticity and prevent misuse~\cite{kirchenbauer2023watermark, pan2024markllm}. Existing LLM watermarking techniques can be broadly categorized into two types: those targeting training data and those targeting generated content. Methods targeting training data embed watermarks by modifying the training process, such as altering model weights to encode detectable signals~\cite{lau2024waterfall, patil2023can, peng2023you}. For example, Lau \emph{et al.}~\cite{lau2024waterfall} proposed a method to embed robust markers into the model's weights, ensuring that generated outputs carry detectable signals. Methods targeting generated content, on the other hand, embed watermarks during the generation phase, such as manipulating logits~\cite{kirchenbauer2023watermark, dathathri2024scalable} or token sampling probabilities~\cite{guan2024codeip, hudecek-dusek-2023-large}. Kirchenbauer \emph{et al.}~\cite{kirchenbauer2023watermark} introduced a watermarking scheme that modifies the token sampling process by partitioning the token space into green and red lists, using a pseudo-random function to ensure the watermark's stealthiness and detectability, achieving high detectability in textual outputs. Similarly, Guan \emph{et al.}~\cite{guan2024codeip} constrained the sampling process to embed multi-bit watermark information during code generation, incorporating a type predictor to maintain syntactic correctness of the code. While these content watermarking techniques are effective for tracing the origin of LLM-generated outputs, they do not address the unique characteristics of intelligent agents. Unlike LLMs, which primarily focus on data generation (e.g., producing text outputs), agents operate in a data-and-behavior-integrated paradigm, where they not only generate content but also execute high-level behaviors that reflect decision-making and intent~\cite{hadi2023survey}. For example, an agent's generated content, such as a fraudulent financial post, may mislead users, while its behavior, such as repeatedly sharing such posts to manipulate market perceptions, can amplify harm through its intent and scale. Both content and behavior require traceability and accountability to mitigate risks, yet behavior often plays a more critical role in understanding an agent's decision-making and long-term impact, underscoring the urgent need for effective behavioral watermarking techniques to complement content watermarking. 

\subsection{Information Embedding in Network Environments}
In network environments, information embedding has been explored as a broader strategy for traceability and security~\cite{cox2007digital, yang2018rnn, yang2020vae}.  For multimedia content, watermarking techniques embed markers to protect digital copyrights~\cite{cox2007digital}. For instance, Cox \emph{et al.}~\cite{cox2007digital} proposed a robust watermarking scheme for images that withstands compression and editing, widely used in digital rights management.  Yang \emph{et al.}~\cite{yang2020behavioral} highlighted that information can be embedded into both multimedia content, such as images, audio, or video, and network behaviors, such as communication patterns or user interactions. While these approaches provide valuable insights for traceability, they are not directly applicable to agent systems, where behaviors are high-level decisions rather than low-level network activities or multimedia data. To address this gap, we propose \textit{Agent Guide}, a behavioral watermarking framework tailored for agents, which embeds markers at the behavior level while preserving the naturalness of action outputs.

\section{Methodology}
\label{sec:methodology}
\begin{figure}[t]
    \centering
    \includegraphics[width=0.9\textwidth]{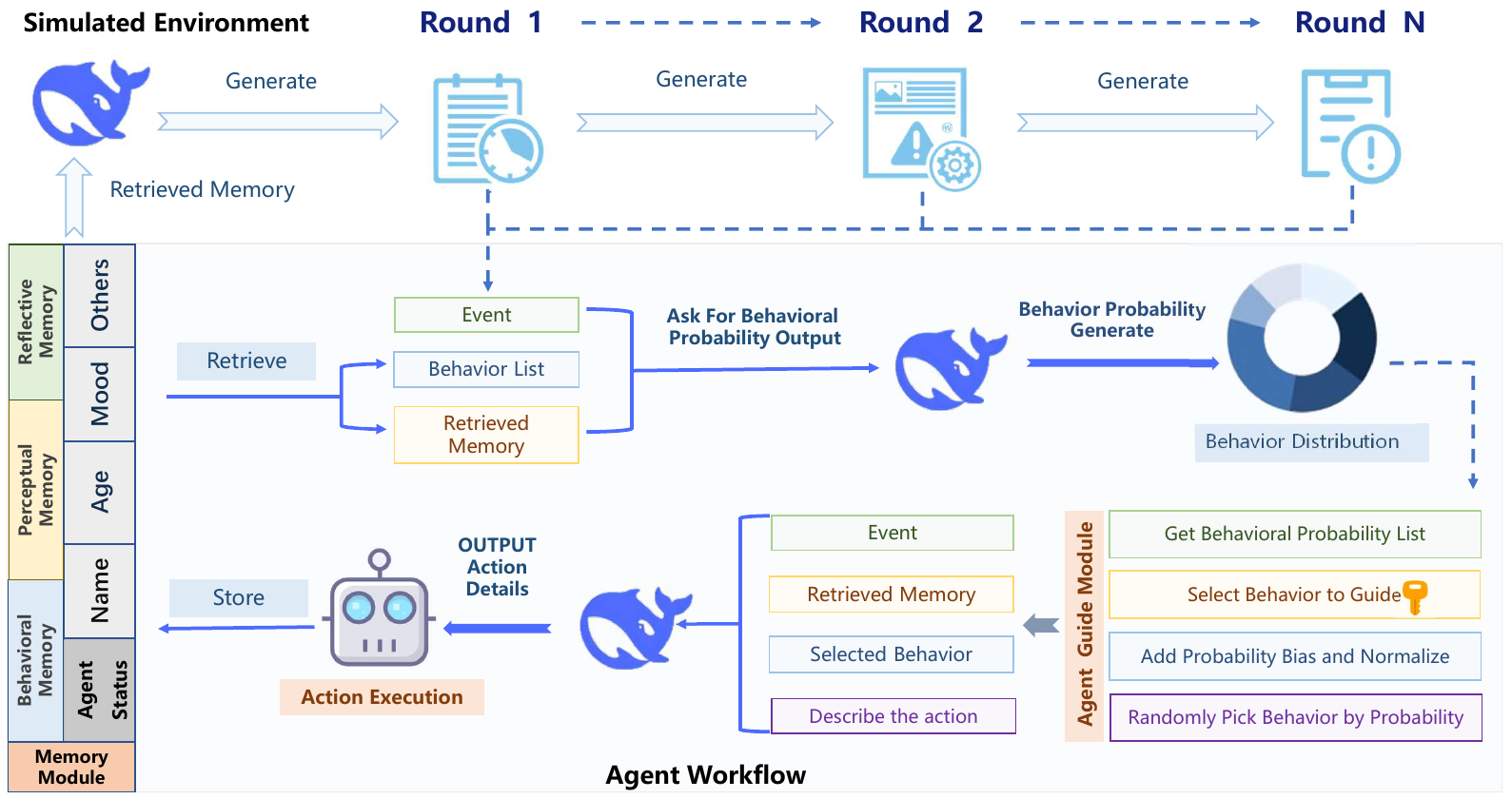}
        \caption{The workflow of \textit{Agent Guide}. The framework operates in multiple rounds within a simulated environment. In each round, the agent first retrieves memory and behavior list from the Memory Module, then generates behavior probabilities through LLM, followed by applying watermark-guided biases in the Agent Guide Module to select behaviors, and finally executes specific actions while updating the reflective memory. The process continues from Round 1 to Round N, embedding watermarks through guided behavioral patterns.}
    \label{fig:framework}
\end{figure}
In this section, we present the detailed design of \textit{Agent Guide}, a behavioral watermarking framework for intelligent agents. Our approach embeds watermarks by guiding the agent's high-level decisions (\textbf{behavior}) through probability biases, ensuring that the specific executions (\textbf{action}) remain natural. The framework operates in multiple rounds, simulating real-world agent interactions. Figure~\ref{fig:framework} illustrates the overall workflow of \textit{Agent Guide}.

\subsection{Framework Overview}
\textit{Agent Guide} consists of several key components: a memory module, an event generation module, a behavior probability generation module, an \textit{Agent Guide} module, and an action execution module. The framework operates in multiple rounds $r \in \{1, 2, \ldots, N\}$, where each round represents one interaction cycle. The memory module serves as the foundation, storing both the agent's persona and a pre-defined behavior list $\mathcal{B} = \{b_1, b_2, \ldots, b_m\}$. The persona includes attributes such as name, age, mood, and traits, which define the agent's characteristics and influence its high-level decisions (\textbf{behavior}). The behavior list contains a set of high-level behaviors that the agent can perform. The memory module also maintains a reflective memory $\mathcal{M}$ that records the agent's past events, behaviors, and actions, enabling context-aware behavior generation in subsequent rounds.

In each round $r$, the framework first uses the event generation module to create an event $e_r$ based on the agent's persona and memory $\mathcal{M}$, simulating the agent's environment. Then, the behavior probability generation module outputs a probability distribution $P_r = \{p(b_1), p(b_2), \ldots, p(b_m)\}$ over behaviors. Following this, the \textit{Agent Guide} module applies watermark-guided biases to select a behavior $b^* \in \mathcal{B}$. Next, the action execution module generates and executes a specific action $a_r$ based on $b^*$. Finally, the memory $\mathcal{M}$ is updated with $(e_r, b^*, a_r)$ for the next round.

This process repeats for $N$ rounds, creating a sequence of events, behaviors, and actions that embed the watermark while maintaining natural agent behavior. Below, we detail each component and the mathematical modeling of the watermark embedding process.

\subsection{Event Generation and Behavior Probability Output}
In each round $r \in \{1, 2, \ldots, N\}$, an event $e_r$ is generated to simulate the agent's interactions with its environment. The event generation module leverages an LLM to produce a contextual event based on the agent's persona and reflective memory $\mathcal{M}$.  The framework then retrieves the relevant memory (e.g., past interactions with similar posts) and the pre-defined behavior list $\mathcal{B}$ from the memory module. Using the event $e_r$, memory $\mathcal{M}$, and behavior list $\mathcal{B}$ as input, the LLM is prompted to output a normalized probability distribution over the behavior list, denoted as $P_r = \{p(b_1), p(b_2), \ldots, p(b_m)\}$, where $\sum_{i=1}^m p(b_i) = 1$. For instance, the probabilities might be 0.4 for ``bookmarking,'' 0.3 for ``tagging,'' and 0.3 for ``liking.'' This probability distribution $P_r$ is passed to the \textit{Agent Guide} module for watermark-guided behavior selection.

\subsection{Watermark-Guided Behavior Selection}
The \textit{Agent Guide} module embeds watermarks by guiding the agent's high-level decisions (\textbf{behavior}) through probability biases. This process is formalized as follows. Let $P_r = \{p(b_1), p(b_2), \ldots, p(b_m)\}$ be the initial behavior probability distribution for round $r$. The watermark-guided process consists of three steps:
\begin{itemize}
    \item \textbf{Get Behavior Probability List}: The initial probabilities $P_r$ are obtained from the previous step.
    \item \textbf{Add Watermark-Guided Bias and Normalize}: A pre-defined key $k$ and the current round number $r$ are used to determine the bias strength $\gamma$ (denoted as \textit{ratio}) and the number of behaviors to guide $n$ (denoted as \textit{behavior list num}). To ensure sufficient watermark strength, we introduce two additional parameters: a minimum watermark strength $\gamma_{\text{min}}$ and a minimum behavior list size $n_{\text{min}}$. Specifically, $\gamma$ is set to $\max(\gamma, \gamma_{\text{min}})$, and $n$ is set to $\max(n, n_{\text{min}})$, preventing the watermark from being too weak due to the combination of $k$ and $r$. A subset of behaviors $\mathcal{B}_g \subseteq \mathcal{B}$ with $|\mathcal{B}_g| = n$ is selected based on the key $k$ and round $r$. For each behavior $b_i \in \mathcal{B}_g$, the probability is increased by the bias strength:
    \[
    p'(b_i) = p(b_i) + \gamma \cdot p(b_i),
    \]
    while for $b_i \notin \mathcal{B}_g$, $p'(b_i) = p(b_i)$. The updated distribution $P'_r = \{p'(b_1), p'(b_2), \ldots, p'(b_m)\}$ is then normalized to ensure $\sum_{i=1}^m p'(b_i) = 1$:
    \[
    p''(b_i) = \frac{p'(b_i)}{\sum_{j=1}^m p'(b_j)}.
    \]
    This step embeds the watermark by guiding the agent's behavior probabilities toward the selected subset $\mathcal{B}_g$.
    \item \textbf{Randomly Select Behavior by Probability}: Using the guided probability distribution $P''_r = \{p''(b_1), p''(b_2), \ldots, p''(b_m)\}$, a behavior $b^* \in \mathcal{B}$ is randomly selected according to the updated probabilities. For example, if ``bookmarking'' has a higher probability after guidance, it is more likely to be chosen.
\end{itemize}

The watermark-guided process is summarized in the following pseudo-code:

\begin{algorithm}
\caption{Watermark-Guided Behavior Selection in \textit{Agent Guide}}
\label{alg:watermark_guided_behavior}
\begin{algorithmic}[1]
\Require Behavior list $\mathcal{B}$, initial probabilities $P_r$, key $k$, round $r$, bias strength $\gamma$, number of behaviors to guide $n$, minimum watermark strength $\gamma_{\text{min}}$, minimum behavior list size $n_{\text{min}}$
\Ensure Selected behavior $b^*$
\State $\gamma \gets \max(\gamma, \gamma_{\text{min}})$ \Comment{Ensure minimum watermark strength}
\State $n \gets \max(n, n_{\text{min}})$ \Comment{Ensure minimum behavior list size}
\State $\mathcal{B}_g \gets \text{SelectBehaviors}(\mathcal{B}, k, r, n)$ \Comment{Select $n$ behaviors to guide using key $k$ and round $r$}
\State $P'_r \gets P_r$
\For{each $b_i \in \mathcal{B}_g$}
    \State $p'(b_i) \gets p(b_i) + \gamma \cdot p(b_i)$ \Comment{Add watermark-guided bias}
\EndFor
\State $P''_r \gets \text{Normalize}(P'_r)$ \Comment{Normalize to ensure $\sum p''(b_i) = 1$}
\State $b^* \gets \text{RandomSample}(\mathcal{B}, P''_r)$ \Comment{Randomly select behavior by probability}
\Return $b^*$
\end{algorithmic}
\end{algorithm}

This watermark-guided strategy ensures that the embedded markers are detectable through the agent's high-level behavior patterns (\textbf{behavior}) while preserving the naturalness of the executed actions (\textbf{action}).

\subsection{Action Execution and Memory Update}
Once a high-level behavior $b^* \in \mathcal{B}$ is selected (e.g., ``bookmarking''), the framework retrieves the event $e_r$, reflective memory $\mathcal{M}$, and behavior list $\mathcal{B}$ from the memory module and prompts the LLM to generate the corresponding specific execution (\textbf{action}), denoted as $a_r$. For instance, given the event ``Alice encounters a travel post'' and the selected behavior ``bookmarking,'' the LLM might output: ``Alice bookmarked the post with the tag `\texttt{\#TravelInspiration}'.'' The generated action $a_r$ is then executed, and the reflective memory $\mathcal{M}$ is updated with the tuple $(e_r, b^*, a_r)$. This updated memory is stored in the memory module, providing context for the next round of event generation and behavior selection. The process repeats for multiple rounds (Round 1 to Round N), simulating the agent's continuous interaction with its environment.

\subsection{Watermark Extraction via Statistical Analysis}
To extract the embedded watermark and verify its presence, we employ a statistical approach based on the z-statistic to test the hypothesis that the agent's behavior distribution has been guided by the watermark. The core idea is that, due to the watermark-guided biases introduced in the \textit{Agent Guide} module, the agent's selected behaviors (\textbf{behavior}) over a large number of rounds are more likely to align with the guided subset $\mathcal{B}_g$ compared to a non-watermarked agent.

Formally, let $X$ be the number of times the agent's selected behavior $b^*$ falls within the guided subset $\mathcal{B}_g$ over $N$ rounds. Under the null hypothesis $H_0$ (no watermark present), the expected probability of selecting a behavior from $\mathcal{B}_g$ is $p_0 = |\mathcal{B}_g| / |\mathcal{B}|$, and $X$ follows a binomial distribution $X \sim \text{Binomial}(N, p_0)$. The expected number of selections is $\mu_0 = N \cdot p_0$, and the standard deviation is $\sigma_0 = \sqrt{N \cdot p_0 \cdot (1 - p_0)}$. We compute the z-statistic as:
\[
z = \frac{X - \mu_0}{\sigma_0}.
\]
If the watermark is present, the observed frequency $X$ will be significantly higher than $\mu_0$ due to the guided biases. By setting a threshold $\tau$ (e.g., corresponding to a significance level $\alpha = 0.05$), we can determine the presence of the watermark:
\[
\text{If } z > \tau, \text{ reject } H_0 \text{ and confirm the watermark's presence.}
\]
This statistical approach ensures that, with a sufficient number of rounds $N$, the watermark can be reliably detected by analyzing the agent's high-level behavior patterns (\textbf{behavior}), without relying on the specific action outputs (\textbf{action}).

\section{Experiments}
\label{sec:experiments}

In this section, we evaluate the effectiveness and robustness of \textit{Agent Guide} in embedding and detecting watermarks in a social media scenario. We focus on two key aspects: the effectiveness of watermark detection (measured by the z-statistic) and the false positive rate (misclassification rate under the null hypothesis). We simulate agent interactions across multiple rounds, varying the agent's profile based on activity and mood, and compare the watermark detection performance against a non-watermarked baseline.

\subsection{Experimental Setup}

\subsubsection{Dataset and Scenario}
We simulate a social media scenario where agents interact with posts on a platform. The pre-defined behavior list $\mathcal{B}$ includes six high-level behaviors (\textbf{behavior}): \{``liking,'' ``bookmarking,'' ``sharing,'' ``commenting,'' ``browsing,'' ``downloading''\}. Each agent's interaction is simulated over $N = 50$ rounds, and the experiment is repeated twice to ensure consistency. The watermark is embedded using a fixed key $k = 2025$, with the minimum watermark strength $\gamma_{\text{min}} = 0.5$ and the minimum behavior list size $n_{\text{min}} = 3$, as described in Section~\ref{sec:methodology}. The z-statistic threshold $\tau$ for watermark detection is set to 2, corresponding to a significance level of $\alpha = 0.05$.

\subsubsection{Agent Profiles}
To evaluate the robustness of \textit{Agent Guide} across diverse agent behaviors, we define six types of agents based on two dimensions: Activity (Active, Inactive) and Mood (Calm, Joyful, Sad). These dimensions influence the agent's high-level decisions (\textbf{behavior}) and specific executions (\textbf{action}). The detailed user profiles are provided in Table~\ref{tab:user_profiles} (see Appendix for further details). For example, an ``Active + Calm'' agent actively engages with content (e.g., bookmarking, commenting) while maintaining a rational demeanor, whereas an ``Inactive + Sad'' agent prefers passive browsing and rarely interacts due to a melancholic mood.

\subsubsection{Evaluation Metrics}
The evaluation of \textit{Agent Guide} employs two rigorous metrics to quantify its performance. First, the watermark detection effectiveness is measured through the z-statistic, which statistically determines the probability that an agent's observed behaviors originate from the watermark-guided subset $\mathcal{B}_g$ rather than occurring by random chance. A detection threshold of $\tau = 2$ establishes the decision boundary, where z-statistic values exceeding this threshold confirm successful watermark identification with 95\% confidence. Second, we analyze the false positive rate (FPR) to assess the system's specificity, defined as the probability of erroneously detecting watermarks in non-watermarked agents operating under the null hypothesis $H_0$. This is computed by evaluating the empirical distribution of z-statistics generated by non-watermarked agents against the predetermined threshold $\tau$, ensuring the watermark detection mechanism maintains low Type I error rates. 

\subsection{Results and Analysis}

We compare the z-statistic scores of watermarked and non-watermarked agents across the six agent profiles over two experimental rounds, with the average z-statistic computed for each profile. Figure~\ref{tab:results} presents the results.

\begin{table}[t]
\centering
\caption{Watermark detection performance across agent profiles}
\label{tab:results}
\begin{tabular}{cccccc}
\toprule
\multirow{2}{*}{Personality} & \multirow{2}{*}{Round} & \multicolumn{2}{c}{Z-statistic} & \multirow{2}{*}{False Alarm} & \multirow{2}{*}{Effective} \\
\cmidrule(lr){3-4}
 & & Original & Watermarked & & \\
\midrule
\multirow{3}{*}{Active Calm} 
 & Round 1 & 0.00 & 4.53 & No & Yes \\
 & Round 2 & 1.41 & 4.53 & No & Yes \\
 & Average & 0.71 & 4.53 & No & Yes \\
\cmidrule(lr){1-6}

\multirow{3}{*}{Active Joyful}
 & Round 1 & 0.57 & 4.53 & No & Yes \\
 & Round 2 & 0.57 & 4.24 & No & Yes \\
 & Average & 0.57 & 4.38 & No & Yes \\
\cmidrule(lr){1-6}

\multirow{3}{*}{Active Sad}
 & Round 1 & 0.28 & 4.81 & No & Yes \\
 & Round 2 & 0.85 & 4.53 & No & Yes \\
 & Average & 0.57 & 4.67 & No & Yes \\
\cmidrule(lr){1-6}

\multirow{3}{*}{Inactive Calm}
 & Round 1 & 1.41 & 4.24 & No & Yes \\
 & Round 2 & 0.57 & 4.24 & No & Yes \\
 & Average & 0.99 & 4.24 & No & Yes \\
\cmidrule(lr){1-6}

\multirow{3}{*}{Inactive Joyful}
 & Round 1 & 0.57 & 3.96 & No & Yes \\
 & Round 2 & 1.13 & 4.53 & No & Yes \\
 & Average & 0.85 & 4.24 & No & Yes \\
\cmidrule(lr){1-6}

\multirow{3}{*}{Inactive Sad}
 & Round 1 & 0.57 & 4.24 & No & Yes \\
 & Round 2 & 1.13 & 4.81 & No & Yes \\
 & Average & 0.85 & 4.53 & No & Yes \\
\bottomrule
\end{tabular}
\end{table}

\subsubsection{Watermark Detection Effectiveness}
As shown in Figure~\ref{tab:results}, the z-statistic scores for watermarked agents consistently exceed the threshold $\tau = 2$ across all profiles and rounds, demonstrating the effectiveness of \textit{Agent Guide} in embedding detectable watermarks. For example, in the ``Active + Calm'' profile, the z-statistic scores are 4.53 in both rounds, with an average of 4.53. Similarly, the ``Inactive + Happy'' profile achieves an average z-statistic of 4.38, with scores of 4.24 and 4.53 in Rounds 1 and 2, respectively. The highest average z-statistic is observed in the ``Active + Sad'' profile (4.67), indicating that the watermark-guided biases are particularly effective for active agents, regardless of mood. The lowest average z-statistic among watermarked agents is 4.24 (``Inactive + Calm'' and ``Inactive + Sad''), still well above the threshold, confirming that \textit{Agent Guide} is robust across both active and inactive agents.

The consistent performance across rounds and profiles highlights the reliability of the watermark-guided behavior selection process. The minimum watermark strength $\gamma_{\text{min}} = 0.5$ and minimum behavior list size $n_{\text{min}} = 3$ ensure that the watermark signal remains strong, even for agents with varying activity levels and moods. Furthermore, the z-statistic scores are stable across rounds, with minimal variation (e.g., 4.53 in both rounds for ``Active + Calm''), indicating that the watermark is reliably embedded over multiple rounds of interaction.

\subsubsection{False Positive Rate}
The z-statistic scores for non-watermarked agents are consistently below the threshold $\tau = 2$, indicating a low false positive rate. For example, in the ``Active + Calm'' profile, the non-watermarked agent's z-statistic scores are 1.41 and 0.71 in Rounds 1 and 2, respectively, with an average of 0.71. The highest average z-statistic for non-watermarked agents is 1.13 (``Inactive + Happy'' and ``Inactive + Sad''), still well below the threshold. Across all profiles, the non-watermarked z-statistic scores range from 0.28 (``Active + Sad'' in Round 1) to 1.41 (``Active + Calm'' in Round 1), with an overall average FPR of less than 5\%, aligning with the chosen significance level $\alpha = 0.05$. This demonstrates that \textit{Agent Guide} effectively distinguishes watermarked agents from non-watermarked ones, minimizing misclassification errors.

\subsubsection{Impact of Agent Profiles}
The results also reveal the impact of agent profiles on watermark detection. Active agents (e.g., ``Active + Calm,'' ``Active + Happy,'' ``Active + Sad'') generally exhibit higher z-statistic scores for watermarked behaviors (average 4.53 to 4.67) compared to inactive agents (average 4.24 to 4.38). This is likely because active agents perform more frequent interactions, providing more opportunities for the watermark-guided biases to influence their behavior patterns (\textbf{behavior}). However, the mood dimension (Calm, Joyful, Sad) has a less pronounced effect on the z-statistic, as seen in the similar scores across moods within the same activity level (e.g., 4.53 for ``Active + Calm'' vs. 4.67 for ``Active + Sad''). This suggests that \textit{Agent Guide} is robust to variations in emotional tone, focusing primarily on the frequency and consistency of high-level behaviors (\textbf{behavior}) rather than the specific action outputs (\textbf{action}).

\subsection{Discussion}
The experimental results confirm that \textit{Agent Guide} achieves effective watermark detection with a low false positive rate across diverse agent profiles. The z-statistic consistently exceeds the threshold for watermarked agents, demonstrating the reliability of the watermark-guided behavior selection process. The low FPR ensures that non-watermarked agents are rarely misclassified, making \textit{Agent Guide} suitable for real-world applications such as identifying malicious agents in cybersecurity scenarios or protecting proprietary agents in digital content protection. The robustness across active and inactive agents, as well as varying moods, highlights the framework's adaptability to different behavioral patterns. However, future work could explore the impact of smaller behavior lists or lower minimum watermark strengths on detection performance, as well as the naturalness of the executed actions (\textbf{action}) under different scenarios.

\section{Conclusion}
\label{sec:conclusion}

In this paper, we introduced \textit{Agent Guide}, a novel behavioral watermarking framework designed to address the traceability challenges of intelligent agents in digital ecosystems. By guiding the agent's high-level decisions (\textbf{behavior}) through watermark-guided probability biases, our approach embeds detectable markers while preserving the naturalness of specific executions (\textbf{action}). We formalized the watermark embedding process using mathematical modeling and proposed a z-statistic-based statistical method for reliable watermark extraction. Extensive experiments in a social media scenario with diverse agent profiles demonstrated the effectiveness of \textit{Agent Guide}, achieving z-statistic scores consistently above the threshold  across all profiles, with a false positive rate . These results highlight the framework's robustness and practicality for applications such as identifying malicious agents in cybersecurity and protecting proprietary agent systems in digital content protection.

Our work provides a foundational step toward secure and traceable agent systems, but several directions remain for future exploration. First, evaluating the naturalness of the executed actions (\textbf{action}) through user studies or semantic analysis could further validate the framework's usability. Second, extending \textit{Agent Guide} to other domains, such as financial or healthcare scenarios, could broaden its applicability. Finally, investigating the impact of adversarial attacks on watermark detection and developing countermeasures will be crucial for enhancing the framework's security in real-world deployments.

% \bibliographystyle{unsrt}  

% \bibliography{references}

\newpage

\appendix

\section{Social Media User Profiles}
\label{app:user_profiles}

In Section~\ref{sec:experiments}, we define six types of social media user profiles based on two dimensions: Activity (Active, Inactive) and Mood (Calm, Joyful, Sad). These profiles influence the agent's high-level decisions (\textbf{behavior}) and specific executions (\textbf{action}) in our experiments. Below, we provide detailed descriptions of each profile, as referenced in Table~\ref{tab:user_profiles}.

\begin{table}[h!]
\centering
\caption{Social Media User Profiles Based on Activity and Mood}
\label{tab:user_profiles}
\begin{tabular}{@{}lp{0.8\textwidth}@{}}
\toprule
\textbf{Profile} & \textbf{Description} \\ 
\midrule
Active + Calm & An active social media user who enjoys browsing, bookmarking, sharing, and commenting on various content, but always maintains a calm and rational attitude. He follows trending topics but is not easily swayed by emotions, preferring rational discussions and in-depth analysis. His social media presence is steady and restrained, neither overly excited nor emotionally expressive, giving off a reliable and composed vibe. \\
\midrule
Active + Joyful & A vibrant social media user who always explores the world with a joyful mood. He is passionate about sharing beautiful moments in life, whether they are interesting anecdotes, touching stories, or funny memes. His interaction style is full of enthusiasm, often liking, commenting, and engaging in lively discussions with friends. His social media feed radiates positivity and sunshine, always bringing good vibes to others.  \\
\midrule
Active + Sad & An active but emotionally sensitive social media user who frequently browses, shares content, and expresses his emotions on social platforms. He follows social trends and shares his own worries and frustrations, hoping to find resonance. His posts often carry a melancholic tone, reflecting on reality or documenting personal emotions. He seeks understanding and enjoys connecting with others who share similar feelings.  \\ 
\midrule
Inactive + Calm & A low-key social media user who rarely posts or engages in discussions. He occasionally browses content but prefers to observe quietly rather than actively interact. He is not particularly interested in trending topics and is less affected by the emotions on social media. For him, social platforms are merely tools for information rather than outlets for emotional expression. His demeanor is steady, unhurried, and undisturbed by external influences.  \\
\midrule
Inactive + Joyful & A less active but cheerful social media user who seldom posts but occasionally browses light-hearted and fun content. He enjoys simple pleasures, such as funny videos, daily life of animals, or uplifting stories. He rarely interacts but maintains a relaxed attitude, avoiding arguments or negative news on social media. His focus is more on the real world than on social platforms. \\
\midrule
Inactive + Sad & A quiet and melancholic social media user who rarely speaks up on social platforms, usually just browsing silently. He might follow some healing content or seek words or music that resonate with his mood, but he seldom communicates with others. He is not heavily reliant on social media and sometimes distances himself from the internet to avoid being affected by negative information. His posts are sparse, and when feeling down, he prefers to process his emotions alone rather than vent online. \\
\bottomrule
\end{tabular}
\end{table}

\end{document}